\documentclass[11pt]{article}
\usepackage{graphicx}
\usepackage{amssymb}
\usepackage{amscd}
\usepackage{mathrsfs}
\usepackage{longtable,lscape}
\usepackage{amsthm}
\usepackage{amsfonts}
\usepackage{amsmath}
\usepackage{bbm}
\usepackage{float}
\usepackage{url}

\newcommand{\captionfonts}{\footnotesize}
\makeatletter  
\long\def\@makecaption#1#2{%
  \vskip\abovecaptionskip
  \sbox\@tempboxa{{\captionfonts #1: #2}}%
  \ifdim \wd\@tempboxa >\hsize
    {\captionfonts #1: #2\par}
  \else
    \hbox to\hsize{\hfil\box\@tempboxa\hfil}%
  \fi
  \vskip\belowcaptionskip}
\makeatother 
\begin{document}
\title{Context and Interference Effects in the Combinations \\ of Natural Concepts}
\author{Diederik Aerts$^1$, Jonito Aerts Argu\"elles$^2$, Lester Beltran$^3$, Lyneth Beltran$^1$, \\ Massimiliano Sassoli de Bianchi$^{4}$, Sandro Sozzo$^{5}$  and Tomas Veloz$^1$ \vspace{0.5 cm} \\ 
        \normalsize\itshape
        $^1$ Center Leo Apostel for Interdisciplinary Studies, 
         Brussels Free University \\ 
        \normalsize\itshape
         Krijgskundestraat 33, 1160 Brussels, Belgium \\
        \normalsize
        E-Mails: \url{diraerts@vub.ac.be},\\\url{lyneth.benedictinelawcenter@gmail.com},\\\url{tveloz@gmail.com}
          \vspace{0.5 cm} \\ 
        \normalsize\itshape
        $^2$ KASK and Conservatory, \\
        \normalsize\itshape
         Jozef Kluyskensstraat 2, 9000 Ghent, Belgium
        \\
        \normalsize
        E-Mail: \url{jonitoarguelles@gmail.com}
	  \vspace{0.5 cm} \\ 
        \normalsize\itshape
        $^3$ 825-C Tayuman Street, \\
         \normalsize\itshape
        Tondo, Manila, The Philippines
         \\
        \normalsize
        E-Mail: \url{lestercc21@gmail.com}
	  \vspace{0.5 cm} \\ 
        \normalsize\itshape
        $^4$ Laboratorio di Autoricerca di Base \\
        \normalsize\itshape
        6914 Lugano, Switzerland
        \\
        \normalsize
        E-Mail: \url{autoricerca@gmail.com}
          \vspace{0.5 cm} \\ 
        \normalsize\itshape
        $^5$ School of Management and IQSCS, University of Leicester \\ 
        \normalsize\itshape
         University Road, LE1 7RH Leicester, United Kingdom \\
        \normalsize
        E-Mail: \url{ss831@le.ac.uk} 
       	\\
              }
\date{}
\maketitle
\begin{abstract}
\noindent
The mathematical formalism of quantum theory exhibits significant effectiveness when applied to cognitive phenomena that have resisted traditional (set theoretical) modeling. Relying on a decade of research on the operational foundations of micro-physical and conceptual entities, we present a theoretical framework for the representation of concepts and their conjunctions and disjunctions that uses the quantum formalism. This framework provides a unified solution to the `conceptual combinations problem' of cognitive psychology, explaining the observed deviations from classical (Boolean, fuzzy set and Kolmogorovian) structures in terms of genuine quantum effects. In particular, natural concepts `interfere' when they combine to form more complex conceptual entities, and they also exhibit a `quantum-type context-dependence', which are responsible of the `over- and under-extension' that are systematically observed in experiments on membership judgments.
\end{abstract}
\section{Introduction\label{intro}}
Philosophers and psychologists have always been interested in the deep nature of human concepts, how they are formed, how they 
combine to create more complex conceptual structures, as expressed by sentences and texts, and how meaning is created in these processes. Unveiling aspects of these mysteries is bound to have a massive impact on a variety of domains, from knowledge representation to natural language processing, machine learning and artificial intelligence. 

The original idea of a concept as a `container of objects', called `instantiations', which can be traced back to Aristotle, was challenged by the first cognitive tests by Eleanor Rosch, which revealed that concepts exhibit aspects, like `context-dependence', `vagueness' and `graded typicality', that prevent a too na\"ive definition of a concept as a `set of defining properties that are either possessed or not possessed by individual exemplars' \cite{r1973,r1978}. More, these tests infused the suspicion that concepts do not combine by following the algebraic rules of classical logic. A first attempt to preserve a set theoretical modeling came from the `fuzzy set approach': concepts would be represented by fuzzy sets, while their conjunction (disjunction) satisfies the `minimum (maximum) rule of fuzzy set conjunction (disjunction)' \cite{z1982}. However, also this approach was confuted by a whole set of experiments by cognitive psychologists, including Osherson and Smith, who identified the `Guppy effect' (or `Pet-Fish problem') in typicality judgments \cite{os1981}, James Hampton, who discovered `overextension' and `underextension' effects in membership judgments \cite{h1988a,h1988b}, and Alxatib and Pelletier, who detected `borderline contradictions' in simple propositions of the form ``John is tall and John is not tall'' \cite{ap2011}. More recently, some of us proved that these data violate Kolmogorov's axioms of classical probability theory \cite{k1933}, thus revealing that classical structures,\footnote{In this paper, we refer to set theoretical structures as `classical structures', because they were originally used to represent systems and interactions in classical physics, and later were 
extended to psychology, economics, statistics, finance, etc. Analogously, we refer to deviations from set theoretical modeling as `deviations from classicality'.} like Boolean and fuzzy set logic and Kolmogorovian probability, are intrinsically unable to model the way in which concepts combine \cite{a2009,ags2013} (Sect. \ref{combinationproblem}).

Interestingly, the deviations from classicality observed in typicality and membership judgments were also identified in other domains of cognitive psychology, and are known as `fallacies of human reasoning', which include conjunctive and disjunctive fallacies, disjunction effects, question order effects, violations of utility theory, etc. \cite{bb2012}. Moreover, in the last decade a novel research programme has taken off, which successfully applies the mathematical formalism of quantum theory (and its possible natural generalizations) to model these fallacies of human reasoning (see, e.g., \cite{a2009,ags2013,bb2012,abgs2013,pb2013,hk2013,pnas2014,asdbs2016}). Even more interestingly, quantum micro-physical entities (like electrons, protons, atoms, etc.) and conceptual entities do exhibit a very similar behavior with respect to `potentiality' and `context-dependence', that is, in both micro-physical and conceptual realms a context is able to change the state of the entity under study, thus actualizing potential properties, rather than just unveiling existing, though unknown, values of them \cite{aa1995}. 

Taking inspiration from our investigations on the operational and realistic approaches to the foundations of quantum physics (see, e.g., \cite{bc81,a1999}), we aim to present in this paper a quantum theoretical framework to represent the conjunction and disjunction of two natural concepts. To this end, we firstly provide an operational-realistic foundation of a theory of concepts, which are defined as `entities in context-dependent states', rather than mere `containers of instantiations' (Sect.~\ref{foundations}). Secondly, we observe that this operational-realistic foundation 
is compatible with the operational-realistic foundation that justifies the use of the Hilbert space formalism to represent the micro-physical entities, which suggests that the mathematical formalism of quantum theory in Hilbert space is a possible (and in a sense natural) candidate to represent conceptual entities too. 

After a brief review of the essentials of the quantum formalism that are needed to understand our results (Sec.~\ref{quantummathematics}), we then introduce the quantum theoretical representation of the conjunction and the disjunction of two natural concepts (Sect.~\ref{context}). In it, deviations from classicality are explained as due to genuine quantum aspects, namely, `context-dependence', `emergence', `interference' and `superposition'. The approach also explains how new conceptual structures 
can emerge when concepts combine. We also show that the quantum theoretical framework is powerful enough to model any type of effect that can be detected in concrete experiments, which, by way of example, we do explicitly with some of Hampton's data for the conjunction and the disjunction of two natural concepts.

We conclude by emphasizing that the quantum theoretical modeling is not `ad hoc', in the sense that it does not arise from 
a mere modeling of data, but rather it results from the foundational hypothesis that quantum theory (and its possible natural extensions) provides a unified and possibly universal paradigm to represent conceptual entities and their meaning-driven interactions (Sect.~\ref{conclusions}).

\section{The combination problem\label{combinationproblem}}
That concepts exhibit aspects of `context-dependence', `vagueness' and `graded typicality' was already known in the seventies since the investigations of Eleanor Rosch \cite{r1973,r1978}. Her studies challenged the traditional view that concepts are containers of instantiations, together with the implicit assumption that conceptual combinations follow the set theoretical algebraic rules of classical logic. In particular, conceptual gradeness suggested a fuzzy set representation of concepts: for each item $X$, a concept $A$ is associated with a graded membership $\mu(A)$, while the conjunction `$A$ and $B$', respectively the disjunction `$A$ or $B$', of the  concepts $A$ and $B$, should satisfy the `minimum rule of fuzzy set conjunction' $\mu(A \ {\rm and} \ B)=\min\, [\mu(A),\mu(B)]$,
respectively the `maximum rule of fuzzy set disjunction' $\mu(A \ {\rm or} \ B)=\max\, [\mu(A),\mu(B)]$ \cite{z1982}.

In this way, one could still maintain that `concepts can be represented as (fuzzy) sets and combine according to set theoretical rules'. However, a whole set of experimental findings revealed that the latter does not hold, even if the combinations are simple conjunctions or disjunctions of two concepts. This raised the so-called `combination problem', that is, how the combination of two concepts 
should  be expressed in terms of the component concepts.

The first obstacle came from the studies of Osherson and Smith. They observed that, for an item like {\it Guppy}, people rate 
its typicality with respect to the conjunction {\it Pet-Fish} higher than its typicality with respect to {\it Pet} or {\it Fish}, taken  
separately. This is the so-called `Guppy effect' (or `Pet-Fish problem') \cite{os1981}. Thus, the Guppy effect violates the minimum rule of fuzzy set conjunction.

But, the most impressive violation of classicality in concept combination came from the experimental studies of psychologist James Hampton, in the late eighties. In a first experiment, Hampton tested the `membership weight' of various sets of items, e.g., {\it Cuckoo}, {\it Peacock}, {\it Toucan}, {\it Parrot}, {\it Raven}, etc., with respect to pairs  of natural concepts, e.g., {\it Bird} and {\it Pet}, taken individually, and their conjunction {\it Bird and Pet} \cite{h1988a}. In a second experiment, Hampton tested the membership weight of various sets of items,  e.g., {\it Apple}, {\it Broccoli}, {\it Tomato}, {\it Mushrooms}, {\it Almonds}, etc., with respect to pairs of natural concepts, e.g., {\it Fruits} and {\it Vegetables}, again taken individually, and their disjunction {\it Fruits or Vegetables} \cite{h1988b}. More explicitly, participants were asked to rate, for each item $X$, its membership with respect to the concepts $A$, $B$ 
and their conjunction `$A$ and $B$', or disjunction `$A$ or $B$' (depending on the pair of concepts considered). 
Membership was estimated on a 7-point Likert scale, $\{ +3,+2,+1,0,-1,-2,-3 \}$, where the choice $+3$ meant that the item was estimated a `very strong member of the concept', the choice $-3$ meant that the item was estimated a `very strong non-member of the concept', and the choice $0$ meant that the participant had not preference of membership or non-membership of the concept. Membership estimations were then converted into relative frequencies and, in the large number limit, into `normalized membership weights'. 

Hampton identified systematic deviations from the minimum rule of fuzzy set conjunction, as well as systematic deviations from the maximum rule of fuzzy set disjunction. Adopting his terminology, if the membership weight of an item $X$ with respect to the conjunction `$A \ {\rm and} \ B$' of two concepts $A$ and $B$ is higher than the membership weight of $X$ with respect to one concept (both concepts), we say that $X$ is `overextended' (`double overextended') with respect to the conjunction. Similarly, 
if the membership weight of an item $X$ with respect to the disjunction `$A \ {\rm or} \ B$' of two concepts $A$ and $B$ is less than the membership weight of $X$ with respect to one concept (both concepts), we say that $X$ is `underextended' (`double underextended') with respect to the disjunction. 

Further experiments confirmed and deepened the findings above. Hampton found overextension in the conjunction `$A$ and not $B$', where `not $B$' denotes the negation of the natural concept $B$ \cite{h1997}. Alxatib and Pelletier identified `borderline contradictions' in sentences involving `$A$ and not $A$' \cite{ap2011}, and some of us also detected overextension and double overextension by simultaneously testing conceptual conjunctions of the form `$A$ and not $B$', `not $A$ and $B$' and `not $A$ and not $B$' \cite{asv2015,asv2016}. 

In these latter works, it emerged that the observed deviations from classicality are deeper than initially expected, as they depend on the fact that set theoretical structures are too limited to represent conceptual combinations. Indeed, let us consider the membership weights of items with respect to concepts and their conjunctions/disjunctions measured by Hampton  \cite{h1988a,h1988b}. In \cite{a2009,ags2013}, we 
proved that a large part of Hampton's data on conjunctions of two concepts cannot be modeled in a single probability space satisfying the axioms of Kolmogorov \cite{k1933}. For example, the item 
{\it Mint} scored in \cite{h1988a} the membership weight $\mu(A)=0.87$, with respect to the concept  
{\it Food}, $\mu(B)=0.81$, with respect to the concept   
{\it Plant}, and $\mu(A \ {\rm and } \ B)=0.9$, with respect to their conjunction {\it Food and Plant}. Hence, the item \emph{Mint} exhibits  a `double overextension' with respect to the conjunction \emph{Food and Plant} of the concepts \emph{Food} and \emph{Plant}, and no Kolmogorovian probability representation exists for these data. More generally, the membership weights $\mu(A), \mu(B)$ and $\mu(A\ {\rm and}\ B)$ of the item $X$ with respect to concepts $A$, $B$ and their conjunction `$A$ and $B$', respectively, can be represented in a single Kolmogorovian probability space if and only if they satisfy the following inequalities \cite{a2009}:
\begin{equation}
\mu(A\ {\rm and}\ B)-\min\,[\mu(A),\mu(B)] \le 0 \qquad \mu(A) + \mu(B) - \mu(A\ {\rm and}\ B) \le 1
\label{ineq01}
\end{equation}
A violation of the first inequality in (\ref{ineq01}) entails, in particular, that the minimum rule of fuzzy set conjunction does not hold, as in the case of {\it Mint}. 

Similarly, in \cite{a2009,ags2013} it was proved that a large part of Hampton's data on disjunctions of two concepts cannot be modeled in a single Kolmogorovian probability space. For example, the item 
{\it Sunglasses} scored in \cite{h1988b} the membership weight $\mu(A)=0.4$ with respect to the concept  
{\it Sportswear}, $\mu(B)=0.2$ with respect to the concept 
{\it Sports Equipment}, and $\mu(A \ {\rm or} \ B)=0.1$
with respect to their disjunction {\it Sportswear or Sports Equipment}. Thus, the item \emph{Sunglasses} exhibits `double underextension' with respect to the disjunction \emph{Sportswear or Sports Equipment} of the concepts \emph{Sportswear} and \emph{Sports Equipment}, and no Kolmogorovian probability representation exists for these data. More generally, the membership weights $\mu(A), \mu(B)$ and $\mu(A\ {\rm or}\ B)$ of the item $X$ with respect to concepts $A$, $B$ and their disjunction `$A$ or $B$', respectively, can be represented in a single Kolmogorovian probability space if and only if they satisfy the following inequalities \cite{a2009}:
\begin{equation} \label{maxdeviation}
\max[\mu(A),\mu(B)]-\mu(A\ {\rm or}\ B)\le 0 \qquad 0 \le \mu(A)+\mu(B)-\mu(A\ {\rm or}\ B)
\end{equation} 
A violation of the first inequality in (\ref{maxdeviation}) entails, in particular, that the maximum rule of fuzzy set disjunction does not hold, as in the case of {\it Sunglasses}.

The difficulties above reveal that the formation and combination rules of human concepts do not obey the restrictions of classical (fuzzy set) logic and Kolmogorovian probability theory. Hence, the combination problem needs to be approached with a novel and more general research programme.

\section{An operational-realistic foundation\label{foundations}}
The investigations of the quantum mechanical representability of concepts by our group in Brussels can be traced back to our 
previous studies on the axiomatic and operational foundations of quantum physics, the differences between classical and quantum structures and the origins of quantum probability (see, e.g., \cite{a1999}). We recognized that any decision process, e.g., a typicality measurement, or a membership estimation, involves a `transition from potential to actual', in which an outcome is actualized within a set of possible outcomes, as a consequence of a contextual interaction (of a cognitive nature) between the decision-maker and the conceptual situation that is the object of the decision. Thus, human decision processes exhibit a deep analogy 
with what occurs in a quantum measurement process, where the measurement context (of a physical nature) influences the measured quantum particle in a non-deterministic way, actualizing properties that were only potential prior to the measurement. Different form `classical probability', which can only deal with situations of lack of knowledge about actuality, `quantum probability' is able to formalize such `contextually driven actualization of potential'. Thus, it can cope with the intrinsic uncertainty underlying both quantum and conceptual realms \cite{aa1995,ag2005a,ag2005b}.

These preliminary analogies led us to systematically inquire into the most plausible mathematical structures formalizing both the 
micro-physical and conceptual entities. In this respect, the formalism of quantum theory, based on complex Hilbert spaces, 
has always amazed researchers for its impressive effectiveness and predictive power. This inspired a fruitful investigation of the foundations of quantum theory in Hilbert space from physically justified axioms, resting on well defined empirical notions, more directly connected with the operations that are usually performed in a laboratory. Such an operational justification would indeed make the formalism of quantum theory more firmly founded.

A
in Geneva and Brussels developed approach to the foundations of quantum physics is the `State Context Property' (SCoP),
in which any physical entity is expressed in terms of the basic notions of `state', `context' and `property', which arise as a consequence of concrete physical operations on macroscopic apparatuses, like preparations and registrations, performed in spatio-temporal domains, like physical laboratories \cite{a1999}. State transformations, measurements, outcomes, and probabilities can then be expressed in terms of these basic notions. If suitable axioms are imposed on the mathematical structures underlying the SCoP formalism, then the Hilbert space structure of quantum theory emerges as a unique mathematical representation, up to isomorphisms  \cite{bc81}.\footnote{Interestingly, the approach allowed to put into evidence an important shortcoming of the standard quantum formalism: the impossibility of describing experimentally separated entities \cite{a1999}.}

This line of research inspired the operational approaches applying the quantum formalism outside the microscopic domain of quantum physics \cite{aa1995,ag2005a,ag2005b}. In particular, a very similar realistic and operational description can be given for the 
conceptual entities  of the cognitive domain, in the sense that the SCoP formalism can be employed to also formalize conceptual entities in terms of states, contexts, properties, measurements and probabilities of outcomes \cite{asdbs2016}.

Let us consider the empirical phenomenology of cognitive psychology. Like in physics, where laboratories define spatio-temporal domains, we can introduce `psychological laboratories', where cognitive experiments are performed. These experiments are performed on situations that are specifically `prepared' for the experiments, including experimental devices and, for example, structured questionnaires, human participants that interact with the questionnaires in written answers, or with each other, e.g., an interviewer and an interviewed. Whenever empirical data are collected from the responses of several participants, a statistics of the obtained outcomes arises. Starting from these empirical facts, we can identify in our approach entities, states, contexts, measurements, outcomes and probabilities of outcomes, as follows. 

The complex of experimental procedures conceived by the experimenter, the experimental design and setting and the cognitive effect that one wants to analyze, define a conceptual entity $A$, and are usually associated with a preparation procedure of a specific 
state of $A$. Hence, like in physics, the preparation procedure sets the  initial state $p_A$ of the conceptual entity $A$ under study. Let us consider, for example, a questionnaire where a participant is asked to rank on a 7-point Likert scale the membership of a list of items with respect to the concepts 
{\it Fruits}, 
{\it Vegetables} and their conjunction 
{\it Fruits and Vegetables}. The questionnaire defines the states  $p_{Fruits}$, $p_{Vegetables}$ and $p_{Fruits \ and \ Vegetables}$ 
of the conceptual entities {\it Fruits}, {\it Vegetables} and {\it Fruits and Vegetables}, respectively. Although in some cognitive situations the preparation procedure of a conceptual entity is hardly controllable, the state of the conceptual entity, defined by means of such a preparation procedure, can always be considered to be a `state of affairs'. It indeed expresses a `reality of the conceptual entity', in the sense that, once prepared in a given state, such condition is independent of any measurement procedure, and can be equally 
confronted with the different participants in an experiment, leading to outcome data and their statistics, exactly like in physics.\footnote{A difference between psychological and physics laboratories is that in the former each participant works as a distinct measuring apparatus, usually producing a single outcome, whereas in the latter a same apparatus is usually used to produce multiple outcomes; see the discussion in \cite{asdb2015}.}

A context $e$ is an aspect of the experiment that can provoke a change of state of the conceptual entity. For example, the concept 
{\it Juicy} can act as a context for the conceptual entity {\it Fruits}, leading to combined concept {\it Juicy Fruits}, which can then be also interpreted as a state of the conceptual entity {\it Fruits}, and more precisely the state describing the situation where `the fruit is juicy'. A particular type of context is the one introduced by the measurement itself. Indeed, when the cognitive experiment starts, an interaction of a cognitive nature occurs between the conceptual entity $A$ under study and a participant in the experiment, in which the state $p_{A}$ of the conceptual entity $A$ generally changes, being transformed into another state $p$. This cognitive interaction is also formalized by means of a context $e$. For example, if the participant is asked to choose among a list of items, say, {\it Olive}, {\it Almond}, {\it Apple}, etc., the most typical one with respect to 
{\it Fruits}, and the answer is 
{\it Apple}, then the initial state 
$p_{Fruits}$ 
of the conceptual entity {\it Fruits} changes to 
$p_{Apple}$, 
i.e. to the state describing the situation `the fruit is an apple',  as a consequence of the contextual interaction with the participant.

Thus, the change of the state of a conceptual entity due to a context may be either `deterministic', hence in principle predictable under the assumption that the initial state
is known, or `intrinsically probabilistic', in the sense that only the probability $\mu(p,e,p_{A})$ that the  state $p_{A}$ of $A$ changes to the state $p$ is given. 
In the example above on typicality estimations, the typicality of the item {\it Apple} for the concept {\it Fruits} is formalized by means of the transition probability $\mu(p_{Apple},e, p_{Fruits})$, where the context $e$ is the context of the typicality measurement.
More generally, suppose that the membership of an item $X$ is estimated by a given sample of participants, with respect to a concept $A$. The item $X$ acts as a context $e_{X}$ that changes the state $p_A$ 
of the conceptual entity $A$ into a new state $p_{X}$. The decision measurement can then be described as a further context $e$ that changes the state $p_{X}$ into a new state $p$. Hence, the membership weight $\mu(A)$ can be expressed as the product of the transition probabilities 
$\mu(A)=\mu(p_{X},e_{X},p_{A})\mu(p,e,p_{X})=\mu(p,e,p_{X})$, where the last equality follows from the fact that the change $p_{A}\to p_{X}$ is deterministic, so that $\mu(p_{X},e_{X},p_{A})=1$.

We have thus described an approach in which, similarly to the operational-realistic foundation of micro-physical entities, a concept can be understood not a container of instantiations, but as an entity in a well-defined state, which can change under the effects of deterministic and indeterministic contexts. This suggests that the Hilbert-based formalism of quantum theory could be a proper candidate to also represent concepts and their interactions.

\section{Essentials of quantum mathematics\label{quantummathematics}}
We present in this section some basic definitions and results of the mathematical formalism of quantum theory that are needed when the quantum formalism is applied to represent concepts and their combinations. We will be rigorous, without however dwelling on technical details.

When the quantum mechanical formalism is applied for modeling purposes, each considered entity  -- in our case a concept -- is associated with a complex Hilbert space ${\mathcal H}$, that is, a vector space over the field ${\mathbb C}$ of complex numbers, equipped with an inner product $\langle \cdot |  \cdot \rangle$ that maps two vectors $\langle A|$ and $|B\rangle$ onto a complex number $\langle A|B\rangle$. We denote vectors by using the bra-ket notation introduced by Paul Adrien Dirac, one of the pioneers of quantum theory \cite{d1958}. Vectors can be `kets', denoted by $\left| A \right\rangle $, $\left| B \right\rangle$, or `bras', denoted by $\left\langle A \right|$, $\left\langle B \right|$. The inner product between the ket vectors $|A\rangle$ and $|B\rangle$, or the bra-vectors $\langle A|$ and $\langle B|$, is realized by juxtaposing the bra vector $\langle A|$ and the ket vector $|B\rangle$, and $\langle A|B\rangle$ is also called a `bra-ket', and it satisfies the following properties:

(i) $\langle A |  A \rangle \ge 0$;

(ii) $\langle A |  B \rangle=\langle B |  A \rangle^{*}$, where $\langle B |  A \rangle^{*}$ is the complex conjugate of $\langle B |  A \rangle$;

(iii) $\langle A |(z|B\rangle+t|C\rangle)=z\langle A |  B \rangle+t \langle A |  C \rangle $, for $z, t \in {\mathbb C}$,
where the sum vector $z|B\rangle+t|C\rangle$ is called a `superposition' of vectors $|B\rangle$ and $|C\rangle$ in the quantum jargon.

From (ii) and (iii) follows that the 
inner product $\langle \cdot |  \cdot \rangle$ is linear in the ket and anti-linear in the bra, i.e. $(z\langle A|+t\langle B|)|C\rangle=z^{*}\langle A | C\rangle+t^{*}\langle B|C \rangle$.

The `absolute value' of a complex number is defined as the square root of the product of this complex number times its complex conjugate, that is, $|z|=\sqrt{z^{*}z}$. Moreover, a 
A complex number $z$ can either be decomposed into its Cartesian form $z=x+iy$, or into its polar form $z=|z|e^{i\theta}=|z|(\cos\theta+i\sin\theta)$, where $|z|$ denotes the `absolute value' of $z$. 
Hence, one has $|\langle A| B\rangle|=\sqrt{\langle A|B\rangle\langle B|A\rangle}$. We define the `length' of a ket vector $|A\rangle$ as $|| |A\rangle ||=\sqrt{\langle A |A\rangle}$. A vector of unitary length is called a `unit vector'. We say that the ket vectors $|A\rangle$ and $|B\rangle$ are `orthogonal', and write $|A\rangle \perp |B\rangle$, if $\langle A|B\rangle=0$.

We have now introduced the necessary mathematics to state the first modeling rule of quantum theory.

\medskip
\noindent{\it First quantum modeling rule.} A state of an entity modeled by quantum theory (in our case a concept) 
is represented by a unit vector $|A\rangle$, that is, $\langle A|A\rangle=1$.

\medskip
\noindent
We also need to introduce the notion of an orthogonal projection operator $M$, which 
is a linear operator on the Hilbert space, that is, a mapping $M: {\mathcal H} \rightarrow {\mathcal H}, |A\rangle \mapsto M|A\rangle$, 
having the properties of being Hermitian and idempotent. This means that, for every $|A\rangle, |B\rangle \in {\mathcal H}$ and $z, t \in {\mathbb C}$, we have:

(i) $M(z|A\rangle+t|B\rangle)=zM|A\rangle+tM|B\rangle$ (linearity)

(ii) $\langle A|M|B\rangle=\langle B|M|A\rangle^{*}$ (hermiticity)

(iii) $M^2=M$ (idempotency)

The identity operator $\mathbbmss{1}$ maps each vector onto itself and is a trivial orthogonal projection operator. We say that two orthogonal projection operators $M_k$ and $M_l$ are orthogonal operators if each vector contained in the range $M_k({\mathcal H})$ is orthogonal to each vector contained in the range $M_l({\mathcal H})$, and we write $M_k \perp M_l$, in this case. The orthogonality of the projection operators $M_{k}$ and $M_{l}$ can equivalently be expressed as $M_{k}M_{l}=0$, where $0$ is the null operator. A set of orthogonal projection operators $\{M_k\ \ \vert \ k=1,\ldots,n\}$ is called a `spectral family' if all projectors are mutually orthogonal, that is, $M_k \perp M_l$ for $k \not= l$, and their sum is the identity operator, that is, $\sum_{k=1}^nM_k=\mathbbmss{1}$.

We are now in a position to state the second and third modeling rules of quantum theory. 

\medskip
\noindent
{\it Second quantum modeling rule.} A measurable quantity $Q$ of an entity modeled by quantum theory (in our case a concept), 
having a set of possible real values $\{q_k \  \vert \  k=1, \ldots, n\}$, is represented by a spectral family $\{M_k \  \vert \  k=1, \ldots, n\}$ in the following way. If the entity is in a state represented by the unit vector $|A\rangle$, then the probability of obtaining the value $q_k$, $k\in \{1, \ldots, n\}$, in a measurement of the measurable quantity $Q$, is
\begin{equation}\label{bornrule}
\mu_{A}(q_k)=\langle A|M_k|A\rangle=||M_k |A\rangle||^{2}
\end{equation}
This formula for probabilistic assignment is called the `Born rule' in the quantum jargon. 

\medskip
\noindent
{\it Third quantum modeling rule.} If the value $q_k$, $k\in \{1, \ldots, n\}$, is actually obtained in the measurement of a measurable quantity $Q$ on an entity modeled by quantum theory (in our case a concept), 
when the entity is in an initial state represented by the unit vector $|A\rangle$, then the initial state is changed into an outcome state 
represented by the vector
\begin{equation}\label{collapse}
|A_k\rangle=\frac{M_k|A\rangle}{||M_k|A\rangle||}=\frac{M_k|A\rangle}{\sqrt{\langle A|M_k|A\rangle}}
\end{equation}
This change of state is called `collapse',  or `reduction', in the quantum jargon.

The quantum modeling above can be generalized in different ways, by introducing rules to model composite entities, or weakening the rules above to represent more complex situations.\footnote{For instance, more general rules of probabilistic assignment than the Born one seem to be necessary for a complete modeling of question order effects data \cite{asdb2015}.} However, what we have here presented is sufficient for attaining our results in the next sections.

\section{Effects of interference and context\label{context}}
The quantum theoretical framework for conceptual combinations is 
obtained by canonically representing the operational notions of state and state changes, (membership and typicality) measurements and (deterministic and indeterministic) contexts, introduced in Sect. \ref{combinationproblem}, by means of the specific Hilbert space mathematics of the quantum formalism, as summarized in Sect. \ref{quantummathematics}. We limit ourselves here to specifying the quantum theoretical framework for the conjunction and the disjunction of two concepts \cite{a2009,ags2013,a2007}.

Let us start with the disjunction of two concepts. Consider, for example, the item {\it Olive}, whose membership was estimated in \cite{h1988b} with respect to the concepts {\it Fruits}, {\it Vegetables} and their disjunction {\it Fruits or Vegetables}. We make three quantum theoretical hypotheses:

(i) Whenever {\it Fruits} and {\it Vegetables} combine, they superpose and interfere. As a consequence of this superposition and interference, a new concept {\it Fruits or Vegetables} emerges.\footnote{This is similar to the prototypical example of the two-slit experiment, where a genuine interference pattern emerges when both slits are open, which cannot be explained in a compositional way, i.e., by assuming that the quantum entities (for example, photons) always pass through one or the other slit.}

(ii) Whenever the item {\it Olive} is considered, a context effect (specific to the item considered) occurs, which produces a deterministic  
change of state of the conceptual entities {\it Fruits}, {\it Vegetables} and {\it Fruits or Vegetables}. This context effect is different than the one created when a different item, say {\it Apple}, is considered, again with respect to {\it Fruits}, {\it Vegetables} and {\it Fruits or Vegetables}.

(iii) The decision of a participant who estimates the membership of {\it Olive} with respect to {\it Fruits}, {\it Vegetables} and {\it Fruits or Vegetables} is considered as a measurement with two outcomes, `yes' and `no', on the conceptual entities  {\it Fruits}, {\it Vegetables} and {\it Fruits or Vegetables}, respectively.

Coming to the representation, let $A$ and $B$ be two concepts and let the membership of the item $X$ be estimated with respect to $A$, $B$ and their disjunction `$A$ or $B$'. Concepts are operationally described as entities in specific states, thus we represent 
the states of the concepts $A$ and $B$ by the unit vectors $|A\rangle$ and $|B\rangle$, respectively, of a Hilbert space $\cal H$, whereas the state of the concept `$A \ \textrm{or} \ B$' is represented by the normalized superposition $|A \ {\rm or} \ B\rangle={1 \over \sqrt{2}}(|A\rangle+|B\rangle)$. For the sake of simplicity, we assume in the following that $|A\rangle$ and $|B\rangle$ are orthogonal, that is, $\langle A| B\rangle=0$. 

To describe the context effect produced by the specific item $X$, we use an orthogonal projection operator $N$, over the Hilbert space $\cal H$. (We can understand $N$ as projecting onto the subspace of states that are also states of the concept $X$). When applied to 
the unit vectors $|A\rangle$ and $|B\rangle$ and ${1 \over \sqrt{2}}(|A\rangle+|B\rangle)$, the operator $N$ produces the new non-unit 
vectors $N|A\rangle$ and $N|B\rangle$ and $N|A \ {\rm or} \ B\rangle= {1 \over \sqrt{2}}(N|A\rangle+N|B\rangle)$. The transformed states of the concepts $A$, $B$ and `$A$ or $B$', resulting from the context effect related to the item $X$, are 
then represented by the unit vectors $|A_{N}\rangle$, $|B_{N}\rangle$ and $|(A \ {\rm or} \ B)_{N}\rangle$, respectively, obtained by normalizing the projected vectors $N|A\rangle$, $N|B\rangle$ and $N|A \ {\rm or} \ B\rangle$, respectively.
More precisely, using (\ref{collapse}), we get
\begin{equation} \label{Nvectors}
|A_{N}\rangle={N|A\rangle \over ||N|A\rangle||},  \,\,\, |B_{N}\rangle={N|B\rangle \over ||N|B\rangle||}, \,\,\, |(A \ {\rm or} \ B)_{N}\rangle={{1 \over \sqrt{2}}(N|A\rangle+N|B\rangle) \over ||{1 \over \sqrt{2}}(N|A\rangle+N|B\rangle)||}.
\end{equation}

Let us now come to the representation of the decision measurement of a person estimating whether the item $X$ is a member of the 
concepts $A$, $B$ and `$A$ or $B$'. This  corresponds to a measurable quantity with two values, `yes' and `no', and is represented by the spectral family $\{M, \mathbbmss{1}-M \}$, with $M$ an orthogonal projection operator over $\cal H$. By using the Born rule (\ref{bornrule}), the probabilities $\mu(A)$, $\mu(B)$ and  $\mu(A\ {\rm or}\ B)$ 
that $X$ is estimated as a member of the concepts $A$, $B$ and `$A$ or $B$', respectively, 
i.e., its membership weights, are given by the inner products $\mu(A)=\langle A_{N}| M|A_{N}\rangle$, $\mu(B)=\langle B_{N}| M|B_{N}\rangle$ and $\mu(A\ {\rm or}\ B)= \langle (A \ {\rm or} \ B)_{N}| M |(A \ {\rm or} \ B)_{N}\rangle$. By using (\ref{bornrule}) and (\ref{Nvectors}), we thus obtain
\begin{eqnarray}
\mu(A)&=&{\langle A|NMN|A\rangle \over \langle A|N|A\rangle}, \quad \mu(B)={\langle B|NMN|B\rangle \over \langle B|N|B\rangle}\\
\mu(A\ {\rm or}\ B)&=&{(\langle A|+\langle B|)NMN(|A\rangle+|B\rangle) \over (\langle A|+\langle B|)N(|A\rangle+|B\rangle)}\nonumber \\
&=&{\langle A|NMN|A\rangle+\langle B|NMN|B\rangle+2\Re\langle A|NMN|B\rangle \over \langle A|N|A\rangle+\langle B|N|B\rangle+2\Re\langle A|N|B\rangle}
\end{eqnarray}
The real parts $\Re\langle A|NMN|B\rangle$ and $\Re\langle A|N|B\rangle$ are the typical `interference terms' of quantum theory. 

We now assume that the context effect consisting of considering the item $X$ and the decision measurement consisting of choosing in favor or against membership of item $X$ are `compatible'. This is a natural assumption, as both contexts are generated by the same item $X$, and is formalized in quantum theory by requiring the commutativity of the corresponding orthogonal projection operators, that is, $MN=NM$. This entails that $NMN=MNN=MN$, hence
\begin{eqnarray} \label{mu(AorB)original}
\mu(A\ {\rm or}\ B)&=&{\langle A|MN|A\rangle+\langle B|MN|B\rangle+2\Re\langle A|MN|B\rangle \over \langle A|N|A\rangle+\langle B|N|B\rangle+2\Re\langle A|N|B\rangle}
\end{eqnarray}
Using some simple algebra and trigonometry, one can show that (\ref{mu(AorB)original}) reduces to the following expression \cite{a2007}:
\begin{equation} \label{sol01}
\mu(A\ {\rm or}\ B)={n^2\mu(A)+n'^2\mu(B)+2nn'\sqrt{\mu(A)\mu(B)}\cos\phi_{d} \over n^2+n'^2+2nn'\cos\phi_{d}(\sqrt{\mu(A)\mu(B)}-\sqrt{(1-\mu(A))(1-\mu(B)})} 
\end{equation}
where $\phi_{d}$ is the `interference angle for the disjunction', and  $n$, $n'$ and $R$ 
are real parameters such that
\begin{equation}
R=\sqrt{(1-\mu(A))(1-\mu(B))}-\sqrt{\mu(A)\mu(B)}, \qquad
\sqrt{(1-n^2)(1-n'^2)}=nn'|R| \label{sol02bis}
\end{equation}

One shows that (\ref{sol01}) provides a solution for any type of effect that can be experimentally detected, namely, classical data satisfying  (\ref{maxdeviation}), overextension and underextension. In addition, one shows that the simplest Hilbert space 
able to do so is the three-dimensional complex Hilbert space ${\mathbb C}^{3}$, with a suitable choice of the orthogonal projection operators $M$ and $N$ \cite{a2007}. For example, consider the item {\it Refrigerator} with respect to the pair of concepts {\it House Furnishings} and {\it Furniture}, and their disjunction {\it House Furnishings or Furniture}. Hampton found $\mu(A)=0.9$, $\mu(B)=0.7$ and $\mu(A\ {\rm or}\ B)=0.575$, which means that we are in the situation of `double underextension' \cite{h1988b}. Equations (\ref{sol01})--(\ref{sol02bis}) can be solved for $R=-0.6205$, 
$n=0.7331$, $n'=0.8312$ and  $\phi_d=119.3535^\circ$. The concepts {\it House Furnishings} and {\it Furniture}  are instead represented by the unit vectors $|A\rangle=(0.6955, 0.2318, 0.6801)$, $|B\rangle=e^{i119.3535^\circ}(0.6955, -0.4553, -0.5559)$, in the canonical base $\{(1,0,0)$, $(0,1,0)$, $(0,0,1) \}$ of ${\mathbb C}^{3}$ \cite{a2007}.


Coming now to the conjunction of two concepts, the same modeling can be used, \emph{mutatis mutandis}, with the conjunction `$A$ and $B$' still represented by a normalized superposition $|A \ {\rm and }\ B\rangle=\frac{1}{\sqrt{2}}(|A\rangle+|B\rangle)$, with the previous `interference angle for the disjunction' $\phi_{d}$ now replaced in (\ref{sol01}) by an `interference angle for the conjunction' $\phi_{c}$.

Again, one shows that (\ref{sol01})  provides a solution for any type of effect that can be experimentally detected, namely, classical data satisfying  (\ref{ineq01}), overextension and underextension, with the simplest Hilbert space being again  ${\mathbb C}^{3}$ \cite{a2007}. For example, consider the item {\it TV} with respect to the pair of concepts {\it Furniture} and {\it Household Appliances}, and their conjunction {\it Furniture and Household Appliances}. Hampton found $\mu(A)=0.7$, $\mu(B)=0.9$, and $\mu(A\ {\rm and}\ B)=0.925$, which means that we are in the situation of `double overextension' \cite{h1988a}. Equations (\ref{sol01})--(\ref{sol02bis}) 
can be solved for $R=-0.6205$, $n=0.5370$, $n'=0.9301$ and  $\phi_c=66.79^\circ$. The concepts {\it Furniture} and {\it Household Appliances}  are instead represented by the unit vectors $|A\rangle=(0.45, 0.29, 0.84)$, $|B\rangle=e^{i66.79^\circ}(0.88, -0.29, -0.37)$, always  in the canonical base of ${\mathbb C}^{3}$ \cite{a2007}.

This completes the construction of a quantum modeling  for the conjunction and the disjunction of two concepts. It shows how new conceptual structures emerge from the component concepts without any need for logical connections between the latter, and explains deviations from classicality in terms of genuine quantum effects, such us context-dependence, interference and superposition. 

\section{Conclusions\label{conclusions}}
We have presented here a quantum theoretical framework to represent natural concepts and their conjunctions and disjunctions. We have shown that such framework can capture genuine quantum aspects, namely, context-dependence, emergence, interference and superposition, and that these aspects are responsible of the deviations from classical logical and probabilistic structures that are observed in membership judgments on conceptual combinations 
\cite{h1988a,h1988b,ap2011,h1997,asv2015,asv2016}.  Hence, the quantum framework provides a solution of the combination problem and constitutes a faithful model for diverse sets of experimental data. 

It is however important to stress that the quantum models arising from the present approach are not `ad hoc', in the sense that they are not devised to merely fit empirical data. They rather emerge from a `theory based approach', which looks for the most plausible mathematical structures to represent both micro-physical and conceptual realms  \cite{a1999,asdb2015}. Indeed, are the deep analogies between the physical and conceptual domains, in the description of measurement processes, that led us to inquire into the realistic-operational foundations of conceptual entities and their description in terms of states, contexts, measurements, outcomes and probabilities, suggesting that quantum structures are very plausible and natural structures to represent both domains. As such, the quantum models are subject to the technical and epistemological constraints of quantum theory, here meant as a possibly universal, coherent and unified theoretical scheme to represent conceptual entities and their interactions.

To conclude, the quantum theoretical framework can be naturally extended to represent more complex conceptual combinations, like conceptual negation and combinations of several concepts. This extension enables the identification of further quantum aspects in conceptual combinations, e.g., `entanglement' and `quantum-type indistinguishability', together with identification of new non-classical patterns of violation, which go beyond over- and under-extension, but capture deep aspects of concept formation \cite{asv2015,asv2016} and context effects \cite{asdb2015}. However, the presentation of these results would go beyond the scopes and length limits of the present paper.

\end{document}